\documentclass[lettersize,journal]{IEEEtran}

\usepackage{amsmath,amsfonts,amssymb}
\usepackage{array}
\usepackage{booktabs}
\usepackage{multirow}
\usepackage{graphicx}
\usepackage[table]{xcolor}
\usepackage{cite}
\usepackage{url}
\usepackage{microtype}

\definecolor{tablegroup}{RGB}{241,244,247}

\newcommand{\method}{CrossVAR}
\newcommand{\dataset}{YREB-SR}
\newcommand{\rev}[1]{#1}

\begin{document}
\bstctlcite{BSTcontrol}

\title{From Sentinel-2 to 1-m Imagery: Real-World $\times10$ Cross-Sensor Super-Resolution via Visual Autoregressive Modeling}

\author{Tailai Chen\textsuperscript{$\dagger$},
        Yuluxin Fu\textsuperscript{$\dagger$},
        Jie Deng\textsuperscript{$*$},
        Shu Luo,
        Yanan Wu,
        and Danfeng Hong,~\IEEEmembership{Senior Member,~IEEE}%
\thanks{Tailai Chen and Jie Deng are with the College of Water Resources and Intelligence Engineering, China Agricultural University, Beijing 100083, China.}%
\thanks{Tailai Chen, Yuluxin Fu, and Yanan Wu are with the College of Information and Electrical Engineering, China Agricultural University, Beijing 100083, China.}%
\thanks{Tailai Chen is also with the Institute of Automation, Chinese Academy of Sciences, Beijing 100190, China, and the University of Chinese Academy of Sciences, Beijing 100049, China.}%
\thanks{Yuluxin Fu is also with the Aerospace Information Research Institute, Chinese Academy of Sciences, Beijing 100094, China, and the University of Chinese Academy of Sciences, Beijing 100049, China.}%
\thanks{Shu Luo is with the Beijing Key Laboratory of Precision Forestry, Beijing Forestry University, Beijing 100091, China.}%
\thanks{Danfeng Hong is with the School of Automation, Southeast University, Nanjing 210096, China.}%
\thanks{This research was supported by the National Natural Science Foundation of China (Grant No. 32402311).}%
\thanks{Tailai Chen and Yuluxin Fu contributed equally to this work.}%
\thanks{Corresponding author: Jie Deng (djcc@cau.edu.cn).}%
\thanks{This work has been submitted to the IEEE for possible publication. Copyright may be transferred without notice, after which this version may no longer be accessible.}%
}

\markboth{IEEE Transactions on Geoscience and Remote Sensing, under review}%
{Chen \MakeLowercase{\textit{et al.}}: YREB-SR and CrossVAR}

\maketitle

\begin{abstract}
Reconstructing 1-m imagery from Sentinel-2 combines two difficulties.  The input and the reference are observed by different satellites, which differ in spatial resolution, spectral response, radiometry, acquisition time, and local geometry, so no synthetic degradation operator links them.  The required magnification of ten in each spatial dimension also far exceeds the factors of two to four at which remote sensing super-resolution is usually developed.  We study this real-world $\times10$ problem using \dataset, which pairs 2022 Sentinel-2 RGB composites with 2022 Jilin-1 RGB references resampled to a 1-m grid over the Yangtze River Economic Belt.  The benchmark contains 11,559 geographically matched scene pairs, with 6,409, 2,135, and 3,015 assigned to training, validation, and testing, respectively.  We develop \method, a visual autoregressive model that predicts the 1-m latent as a nine-scale hierarchy conditioned on the Sentinel-2 input and estimates the remaining quantization residual with a diffusion refiner run without injected noise.  Complete $1600\times1600$ scenes are reconstructed in overlapping windows that start on the Sentinel-2 pixel grid, as every training patch does, and every window is decoded at the size on which the decoder was trained before the windows are blended with Gaussian weights.  On all 3,015 test scenes, \method\ achieves 27.60\ dB PSNR, 0.7725\ SSIM, 0.1451\ LPIPS, and 7.15\ FID, the best value of each metric among the compared methods.  These results establish a strong autoregressive baseline for large-scale cross-sensor reconstruction. To facilitate reproducibility, the source code and implementation are publicly available at https://github.com/SHANOAL/RSSR.
\end{abstract}


\section{Introduction}
\label{sec:introduction}
\IEEEPARstart{S}{entinel-2} provides frequently, openly available observations for monitoring large geographic areas, but its 10-m visible bands do not resolve many roads, roofs, field boundaries, and small water features \cite{Drusch2012}.  Very-high-resolution imagery can reveal these structures, although its coverage and acquisition conditions differ from those of medium-resolution missions.  Super-resolution (SR) offers a way to estimate finer spatial structure from the available observations.  Estimating 1-m imagery from Sentinel-2 is, however, considerably harder than the SR settings commonly studied in remote sensing, for two reasons.

\begin{figure}[!t]
    \centering
    \includegraphics[width=\columnwidth]{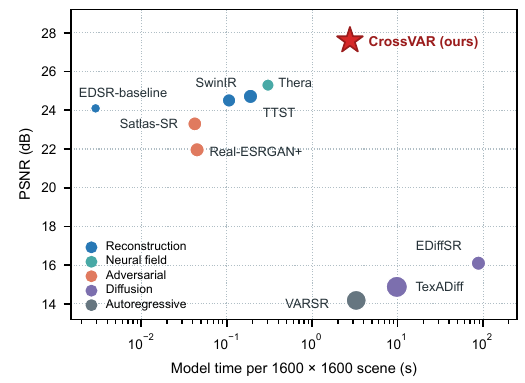}
    \caption{PSNR and model time per complete $1600\times1600$ test scene of \dataset.  Marker area grows with the number of parameters.}
    \label{fig:tradeoff}
\end{figure}

First, the input and the reference are observed by different sensors.  Most remote sensing SR methods learn from pairs in which the low-resolution (LR) image is synthesized from the high-resolution (HR) image by bicubic downsampling or a predefined blur kernel \cite{wang2022rsreview,qi2026rssurvey}.  The two images then share the sensor, the acquisition time, and the radiometric processing, and the degradation is known exactly.  Models trained on such pairs lose accuracy on real acquisitions, because the simulated degradation departs from the real one \cite{wang2025rrsisr}.  Independent satellites differ in optics and modulation transfer function, spectral response, radiometric calibration and product generation, atmospheric state, and acquisition date.  In \dataset, a single three-channel input is a Sentinel-2 composite from 2022, whereas its Jilin-1 reference was acquired in 2022.  Land-cover change and seasonal appearance thus add to the sensor differences.  Their footprints are geographically matched, but exact pixel-wise correspondence is not assumed.  The LR image is therefore not a degraded copy of the reference, and no degradation operator links the two.  A model must learn the cross-sensor relation from paired data without treating every sensor or temporal disagreement with the reference as missing detail.  Real-world benchmarks pair independent acquisitions for this reason, including Sentinel-2 with WorldView-2 in MuS2 \cite{kowaleczko2023mus2}, with SPOT 6/7 in WorldStrat \cite{cornebise2022worldstrat}, and with NAIP aerial imagery in S2-NAIP \cite{wolters2023satlas}, as well as Landsat 8, Sentinel-2, and PlanetScope in L8S2PS \cite{meng2026rsreview}.

Second, the magnification is far larger than in common practice.  Most remote sensing SR methods operate at factors of two to four, because the problem becomes rapidly more ill-posed at higher magnification \cite{shi2024wavediffur}, and real-world benchmarks likewise use moderate factors, such as three in L8S2PS \cite{meng2026rsreview}, or provide several Sentinel-2 acquisitions per reference \cite{kowaleczko2023mus2,cornebise2022worldstrat}.  Reconstructing 1-m imagery from a single 10-m Sentinel-2 image requires $\times10$, so each input pixel is expanded into 100 output pixels, compared with 16 at $\times4$.  The nominal $\times10$ ratio describes the output grid; it does not imply that one observed LR pixel is the arithmetic mean of the corresponding 100 reference pixels.  Most of the structure in a 1-m output, such as individual buildings, narrow roads, and field boundaries, lies below the input sampling and must come from learned image priors.  Strong priors produce sharp detail that the observation does not necessarily support, and the cross-sensor differences make the reference an imperfect target for learning such priors.  The large factor also enlarges every scene, as a $160\times160$ Sentinel-2 input becomes a $1600\times1600$ output that exceeds the crop size used in training, so complete scenes must be assembled from overlapping windows without visible seams.  Reconstruction-oriented CNNs and transformers provide fidelity baselines for this setting \cite{lim2017edsr,liang2021swinir,xiao2024ttst}, and adversarial and diffusion models offer perceptual priors \cite{wang2021realesrgan,xiao2024ediffsr}, so their comparison must use the same complete scenes and assess both reference fidelity and perceptual similarity.

Visual autoregressive modeling (VAR) predicts an image through a coarse-to-fine sequence of latent maps rather than a raster-ordered sequence of individual pixels \cite{tian2024var}, and the hierarchy has also been conditioned on LR images for SR \cite{qu2025varsr}.  The hierarchy suits a large magnification, since its coarse scales fix the scene layout that the Sentinel-2 input constrains, and its finer scales add local structure from the learned prior.  \method\ encodes the Sentinel-2 input into one condition token per latent position, predicts the 1-m latent as nine code maps with a next-scale transformer, and estimates the residual that the codes leave with a residual refiner.  Two steps of this inference add error unrelated to the reference.  First, the residual refiner samples the residual with injected noise, which adds variation that no reference supports.  Second, the assembly of complete scenes departs from the conditions of training.  At $\times4$, a 16-pixel latent position spans exactly four input pixels, so every window on the latent grid also starts on the input grid.  At $\times10$, input pixels span 10 output pixels and latent positions span 16, so the two grids coincide only every 80 output pixels, and overlapping windows mostly start inside an input pixel, whereas every training patch starts on the input grid.  Fusing the windows and decoding the scene in one pass further applies the VQ decoder to a latent field sixteen times larger than those seen in training.  \method\ removes the first error by deterministic residual refinement and the second by aligned window decoding, which starts every window on the input grid and decodes it at the training size.  Both act at inference and leave the trained weights unchanged.  Fig.~\ref{fig:tradeoff} places \method\ among the compared methods by fidelity and cost on the complete test scenes.

We evaluate \method\ on the real paired benchmark described in Section~\ref{sec:dataset} and summarized in Table~\ref{tab:dataset-spec}.  The model processes the natural-color input alone; auxiliary spectral bands, temporal sequences, semantic labels, and geographic metadata are not provided to it.  The contributions of this work are as follows.

\begin{itemize}
    \item \textit{A real cross-sensor $\times10$ benchmark.} \dataset\ contains 11,559 Sentinel-2/Jilin-1 scene pairs with a fixed scene-level partition and documented sampling, geospatial standardization, and quality control.  The test unit is the complete scene, not a selected crop.
    \item \textit{An autoregressive reconstruction of complete scenes.} \method\ predicts the 1-m latent coarse to fine from the Sentinel-2 input, refines the quantization residual without injected noise, and reconstructs $1600\times1600$ scenes by aligned window decoding, which keeps every window on the input grid and at the training size of the decoder.
    \item \textit{A multi-family empirical comparison.} We evaluate reconstruction, continuous-representation, adversarial, diffusion, and autoregressive baselines with common RGB metrics on 3,015 scenes.  We report fidelity, perceptual quality, and efficiency for every method.
\end{itemize}

\section{Related Work}
\label{sec:related}

\subsection{Super-Resolution Datasets}
Most remote sensing SR datasets synthesize the LR image from the HR image by bicubic downsampling or a predefined blur kernel \cite{wang2022rsreview,qi2026rssurvey}.  Real-world datasets instead pair independent acquisitions of the same area.  MuS2 pairs multiple Sentinel-2 revisits with WorldView-2 references \cite{kowaleczko2023mus2}, WorldStrat matches SPOT 6/7 imagery with multiple Sentinel-2 acquisitions \cite{cornebise2022worldstrat}, S2-NAIP pairs Sentinel-2 with NAIP aerial imagery \cite{wolters2023satlas}, and OpenSR-test curates cross-sensor sets for evaluating Sentinel-2 SR \cite{aybar2024opensrtest}.  RRSISR captures LR and HR images of the same scene with a spectral camera \cite{wang2025rrsisr}, and L8S2PS aligns Landsat 8, Sentinel-2, and PlanetScope images at a factor of three \cite{meng2026rsreview}.  \dataset\ pairs a single Sentinel-2 composite with a 1-m Jilin-1 reference at $\times10$ and evaluates complete scenes.

\subsection{Super-Resolution}
Reconstruction networks learn a direct mapping from the LR to the HR image, with residual convolutional blocks in EDSR \cite{lim2017edsr}, window-based transformer blocks in SwinIR \cite{liang2021swinir}, and top-$k$ token selection for remote sensing in TTST \cite{xiao2024ttst}.  Arbitrary-scale methods decouple the output grid from a fixed factor, as LIIF queries a continuous representation from local features \cite{chen2021liif} and Thera integrates a Gaussian point-spread function with neural heat fields \cite{becker2023thera}.  Generative methods pursue perceptually plausible detail \cite{ledig2017srgan,wang2021realesrgan} under the trade-off between distortion and perceptual quality \cite{blau2018perception}.  Satlas-SR applies adversarial training to Sentinel-2 \cite{wolters2023satlas}, diffusion priors are explored by EDiffSR, RSDiffSR, SGDM, and TexADiff \cite{xiao2024ediffsr,zhu2025rsdiffsr,wang2025sgdm,zhang2026texadiff}, SEN2SR emphasizes radiometric and spatial consistency with the input \cite{aybar2026sen2sr}, and WaveDiffUR addresses ultra-large magnification with a wavelet-domain diffusion solver \cite{shi2024wavediffur}.

\subsection{Visual Autoregressive Modeling}
Vector-quantized autoencoders map images to discrete latent codes \cite{oord2017vqvae,esser2021vqgan}, and VAR models these codes by next-scale prediction from coarse to fine \cite{tian2024var}.  VARSR conditions this hierarchy on LR prefix tokens and estimates the quantization residual with a diffusion refiner \cite{qu2025varsr}.  These models are developed on natural images at moderate factors.  \method\ applies next-scale prediction to real Sentinel-2/Jilin-1 pairs at $\times10$, refines the residual without injected noise, and reconstructs complete scenes in windows aligned with the input grid.

\section{YREB-SR Dataset Construction}
\label{sec:dataset}

\subsection{Study Area and Sampling Design}

The Yangtze River Economic Belt (YREB) extends from the mountainous upper reaches of the Yangtze River to the densely urbanized and intensively cultivated middle and lower reaches. The study area encompasses all administrative regions of 11 provincial-level units and relevant county-level divisions in eight additional provinces, comprising 1,173 county-level administrative units and more than 2.4 million~km$^2$ of land \cite{MNR2024,Deng2026JointSeg}. The pronounced variation in topography, climate, settlement density, vegetation, agricultural systems, and urban morphology provides a geographically diverse setting for evaluating real-world remote sensing super-resolution.

\begin{figure*}[!t]
    \centering
    \includegraphics[
        width=\linewidth,
        height=0.95\textheight,
        keepaspectratio
    ]{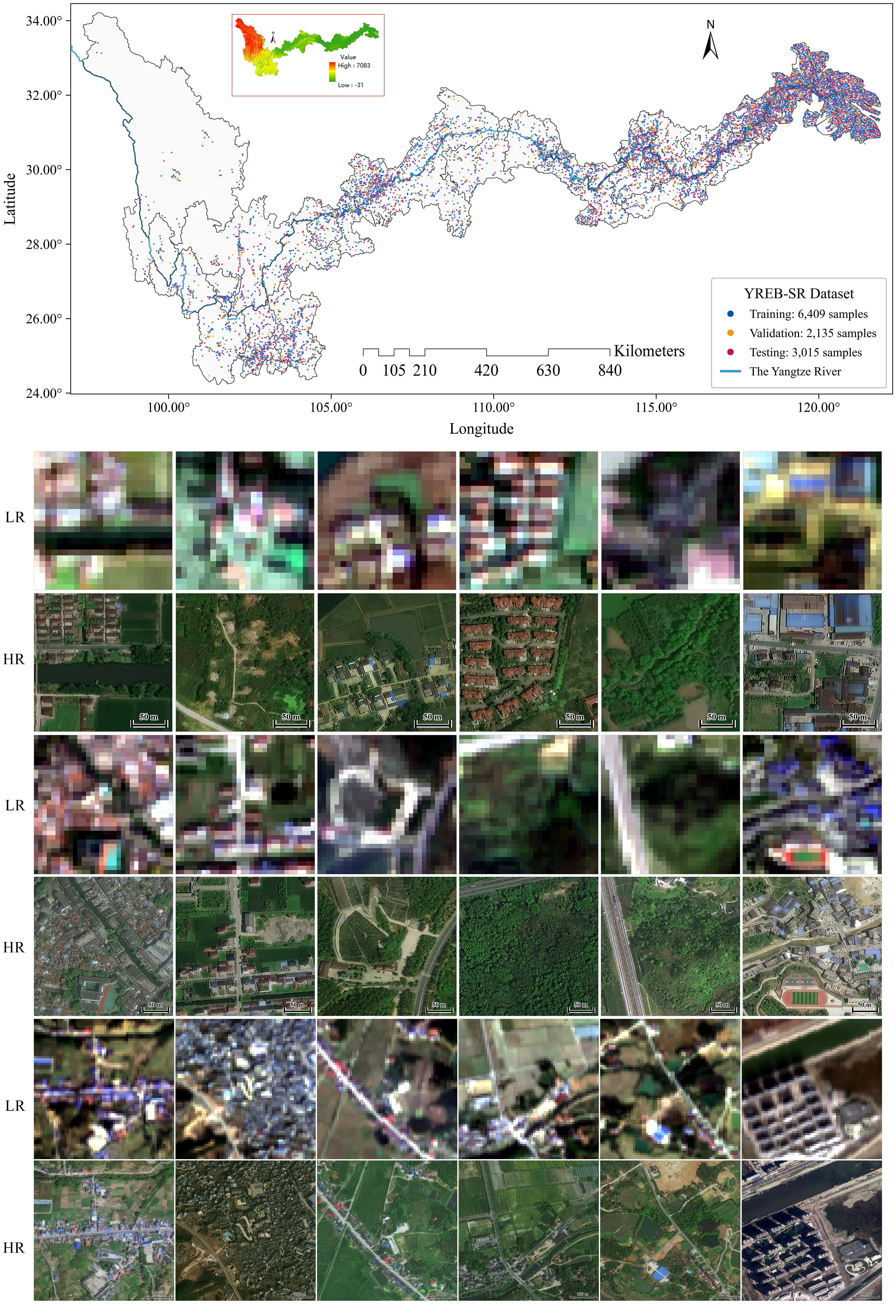}
    \caption{Spatial distribution of the 11,559 cross-sensor scene pairs in \dataset{} across the Yangtze River Economic Belt. Colors indicate the training, validation, and test subsets, while blue lines represent the Yangtze River network.}
    \label{fig:yreb_distribution}
\end{figure*}

Candidate locations were selected using the built-up class of ESA WorldCover as a spatial guide \cite{Zanaga2022WorldCover}. Stratified random sampling was conducted across the upstream, midstream, and downstream portions of the YREB to capture regional differences in surface structure and urbanization intensity. A minimum center-to-center distance of 3~km was imposed between sampling locations. Because each scene covers approximately $1.6\times1.6$~km$^2$, this constraint prevents direct overlap between scene footprints and reduces immediate spatial redundancy. Although the sampling strategy emphasizes built-up environments containing high-frequency structures such as buildings, roads, and fragmented impervious surfaces, the resulting scenes also encompass vegetation, cropland, water bodies, bare land, and mixed rural landscapes. Figure~\ref{fig:yreb_distribution} shows the geographic distribution of the resulting scene pairs and their dataset assignments.

\subsection{Cross-Sensor Image Sources}

\subsubsection{Sentinel-2 Low-Resolution Images}

The low-resolution (LR) images were derived from Sentinel-2 observations accessed through Google Earth Engine \cite{Gorelick2017GEE}. Source acquisitions with reported cloud coverage exceeding 20\% were excluded, after which median compositing was applied to the remaining 2022 observations to reduce residual cloud and atmospheric contamination. Sentinel-2 provides a native spatial resolution of 10~m in the blue (B2), green (B3), and red (B4) bands \cite{Drusch2012}. These bands were retained and ordered as B4--B3--B2 to form natural-color RGB inputs.

For each sampling location, a $160\times160\times3$ LR scene was extracted at the native 10-m grid spacing, corresponding to a nominal ground footprint of $1.6\times1.6$~km$^2$. Each benchmark sample provides a single three-channel Sentinel-2 image as the model input. Other Sentinel-2 bands, multitemporal sequences, Sentinel-1 observations, semantic labels, and geographic metadata are not supplied to the reconstruction models. This design isolates the ability of each method to reconstruct fine spatial detail from natural-color medium-resolution imagery.

\subsubsection{Jilin-1 High-Resolution Reference Images}

The high-resolution (HR) references were derived from 0.75-m RGB imagery acquired in 2022 by the Jilin-1 satellite constellation \cite{Jilin1Data}. The source imagery was resampled to a uniform 1-m grid, from which $1600\times1600\times3$ RGB scenes were extracted. Each HR scene therefore covers the same nominal $1.6\times1.6$~km$^2$ footprint as its corresponding Sentinel-2 input.

The Jilin-1 images are regarded as cross-sensor references rather than strictly simultaneous ground truth. The LR and HR observations differ in acquisition date, sensor optics, modulation transfer function, spectral response, radiometric representation, atmospheric conditions, and product-generation chain. These differences are intentionally retained because \dataset{} is designed to evaluate practical cross-sensor reconstruction rather than inversion of an analytically specified synthetic degradation.

\subsection{Geospatial and Radiometric Standardization}

For each sampling location, the Jilin-1 and Sentinel-2 images were extracted over the same geospatial extent using their associated georeferencing information. The Jilin-1 imagery was resampled by cubic interpolation to a $1600\times1600$ grid at a nominal 1-m ground sampling distance. The complete Sentinel-2 raster was independently reprojected and resampled over the corresponding extent to a $160\times160$ grid before extraction of bands B4, B3, and B2. Only locations for which both modalities were available and readable were retained, resulting in 11,559 complete RGB scene pairs.

All standardized pairs are stored in the WGS~84 geographic coordinate system (EPSG:4326). Within each pair, the LR and HR rasters have identical geographic bounds and resolution-consistent affine transforms. Source projection information, spatial extent, affine transformation parameters, and NoData masks are propagated through the standardization procedure whenever available.

Sentinel-2 digital values are divided by 10,000 and stored as \texttt{float32} surface reflectance, whereas the HR references retain their \texttt{uint8} RGB representation. No histogram matching, color transfer, or sensor-to-sensor radiometric normalization is applied. The benchmark therefore preserves the actual cross-sensor radiometric discrepancy rather than imposing an artificially harmonized color relationship.

The quality-control procedure verifies file readability, paired filenames, band availability, raster dimensions, data types, geographic bounds, affine transforms, and NoData values. Dataset-level streaming statistics are additionally computed for each band, including the minimum, maximum, mean, standard deviation, and value histogram. Missing, corrupted, or unmatched files are recorded in an error log and excluded before dataset partitioning.

Unlike conventional SR datasets, the LR images in \dataset{} are not generated from the HR references using bicubic downsampling, Gaussian filtering, additive noise, or a predefined degradation kernel. Instead, each pair consists of independent observations acquired by two satellite systems. The LR--HR relationship consequently includes sensor-specific optical responses, radiometric differences, temporal and seasonal changes, atmospheric variability, and residual local misregistration. It cannot be represented by a single deterministic degradation operator.

The pairs are geographically matched at the scene level, but exact pixel-wise correspondence is not assumed. A 10-m Sentinel-2 pixel nominally corresponds to a $10\times10$ region on the 1-m reference grid; however, its measured value is determined by the Sentinel-2 imaging process rather than by direct averaging of the corresponding 100 Jilin-1 pixels. This distinction separates \dataset{} from synthetically downsampled benchmarks and more closely reflects operational cross-sensor reconstruction.

\subsection{Dataset Partitioning and Patch Generation}

Following quality control, the 11,559 scenes were randomly shuffled using a fixed seed and partitioned at the scene level into 6,409 training, 2,135 validation, and 3,015 test pairs. These subsets account for 55.45\%, 18.47\%, and 26.08\% of the complete dataset, respectively. Scene identities are mutually exclusive across the three subsets. Furthermore, the 3-km minimum sampling distance ensures that their image footprints do not directly overlap, although the random partition is not intended to constitute a geographically blocked holdout experiment.

Each training and validation scene is divided into a nonoverlapping $4\times4$ grid. This produces aligned $40\times40$ LR and $400\times400$ HR patch pairs while preserving the nominal $\times10$ scale. The procedure yields 102,544 training and 34,160 validation patch pairs. During model training, random horizontal and vertical flips are applied identically to the LR and HR members of each pair to preserve correspondence.

The validation subset is used for model selection and hyperparameter configuration, whereas the fixed test subset is excluded from both parameter estimation and configuration selection. All competing methods use the same official partition. Test predictions are reconstructed and evaluated over all 3,015 complete $1600\times1600$ scenes, without border removal or selection of favorable test crops. The resulting protocol provides identical spatial support for interpolation-based, convolutional, transformer-based, continuous-representation, adversarial, and autoregressive methods.

\begin{table}[!t]
\caption{Summary of the \dataset{} Benchmark}
\label{tab:dataset-spec}
\centering
\setlength{\tabcolsep}{5pt}
\renewcommand{\arraystretch}{1.03}
\begin{tabular}{@{}p{0.34\columnwidth}p{0.58\columnwidth}@{}}
\toprule
Attribute & Setting \\
\midrule
\rowcolor{tablegroup}
\multicolumn{2}{@{}l}{\textit{Cross-sensor image pairs}} \\
Sensor pair & Sentinel-2 RGB $\rightarrow$ 1-m RGB reference \\
Spectral channels & B4--B3--B2 $\rightarrow$ R--G--B \\
Nominal GSD & 10~m $\rightarrow$ 1~m ($\times10$) \\
Full-scene grids & $160\times160 \rightarrow 1600\times1600$ pixels \\
Training grids & $40\times40 \rightarrow 400\times400$ pixels \\
Storage types & \texttt{float32} LR; \texttt{uint8} HR \\
Spatial reference & EPSG:4326 with pairwise matched bounds \\
\addlinespace[2pt]
\rowcolor{tablegroup}
\multicolumn{2}{@{}l}{\textit{Official partition and evaluation}} \\
Full scenes & 6,409 training; 2,135 validation; 3,015 test \\
Aligned patches & 102,544 training; 34,160 validation \\
Test protocol & 3,015 complete scenes without border cropping \\
\bottomrule
\end{tabular}
\end{table}

Overall, \dataset{} differs from commonly used synthetic SR datasets in three principal respects. First, it evaluates an extreme $\times10$ reconstruction from native 10-m Sentinel-2 imagery to a 1-m VHR reference. Second, the LR images are independent cross-sensor observations rather than artificial degradations of the HR targets. Third, evaluation is conducted on a large and geographically heterogeneous collection of complete scenes under a fixed protocol. Together, these characteristics establish a reproducible and practically relevant benchmark for real-world cross-sensor remote sensing super-resolution.

\section{Method}
\label{sec:method}
\begin{figure*}[!t]
    \centering
    \includegraphics[width=\textwidth]{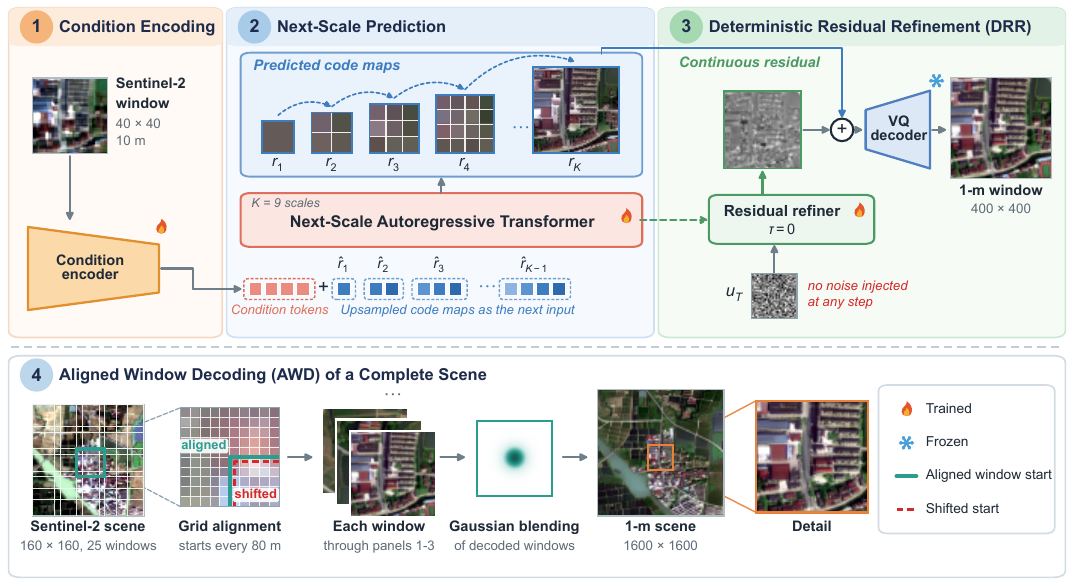}
    \caption{Overview of \method.  Panels 1 to 3 reconstruct one window.  The condition encoder maps the Sentinel-2 window, resized to the output grid, to one condition token per latent position, and the next-scale transformer predicts nine code maps from $1\times1$ to $25\times25$.  Deterministic residual refinement (DRR) estimates the residual of the codes without injected noise, and the frozen VQ decoder outputs the 1-m window.  Aligned window decoding (AWD) in panel 4 starts every window of a complete scene on the input pixel grid, reconstructs it through panels 1 to 3, and blends the decoded windows with Gaussian weights.}
    \label{fig:framework}
\end{figure*}

\method\ reconstructs a 1-m image window by window, and Fig.~\ref{fig:framework} shows its stages.  Section~\ref{sec:prelim} defines the task and the latent representation of a window.  The condition encoder in Section~\ref{sec:cond} maps the Sentinel-2 window to condition tokens, the next-scale transformer in Section~\ref{sec:nsp} predicts nine code maps of the latent from these tokens, and the residual refiner in Section~\ref{sec:drr} estimates the residual that the codes leave.  Section~\ref{sec:objective} gives the training objective, and Section~\ref{sec:wd} reconstructs complete scenes from windows.

\subsection{Preliminaries}
\label{sec:prelim}
Let $\mathbf{x}\in\mathbb{R}^{3\times H\times W}$ denote the standardized Sentinel-2 input and $\mathbf{y}\in[0,1]^{3\times 10H\times 10W}$ its Jilin-1 reference.  We learn a reconstruction $\hat{\mathbf{y}}=F_{\theta}(\mathbf{x})$ on the nominal 1-m grid.  As specified in Section~\ref{sec:dataset}, $\mathbf{y}$ is a geographically matched cross-sensor reference rather than an exactly registered observation, so $F_{\theta}$ is learned from the paired data instead of being derived from a degradation operator.  \method\ operates on windows of $40\times40$ input pixels, the size of the training patches, and reconstructs each window at $400\times400$ pixels.

A VQ autoencoder represents a $400\times400$ window by a latent on a $25\times25$ grid.  Its encoder $E$ reduces each spatial dimension by a factor of 16, and its decoder $G$ maps a latent back to the image.  The reference latent $\mathbf{z}^{*}=E(\mathbf{y})$ is represented by $K=9$ code maps $\mathbf{r}_{1:K}$ drawn from one codebook shared by all scales.  The map $\mathbf{r}_k$ has $s_k\times s_k$ positions with $(s_1,\ldots,s_9)=(1,2,3,4,6,9,13,18,25)$, and the maps are obtained by residual quantization from coarse to fine,
\begin{equation}
    \mathbf{r}_k=Q_k\big(\mathbf{z}^{*}-\mathbf{z}_{<k}\big),\qquad
    \mathbf{z}_{<k}=\sum_{j<k}\phi_j(\mathbf{r}_j),
    \label{eq:rq}
\end{equation}
where $Q_k$ assigns the nearest code to each position of its argument downsampled to $s_k\times s_k$, and $\phi_j$ upsamples the code embeddings $\mathbf{e}(\mathbf{r}_j)$ to the $25\times25$ grid and applies a residual convolution.  The autoencoder stays frozen, and \method\ predicts the latent $\hat{\mathbf{z}}$ of a window, which the decoder maps to the 1-m window $\hat{\mathbf{y}}=G(\hat{\mathbf{z}})$.

\subsection{Condition Encoding}
\label{sec:cond}
The Sentinel-2 window is first resized to the $400\times400$ output grid by bicubic interpolation, so that the condition and the reconstruction share pixel coordinates.  The condition encoder is a convolutional network with four stride-2 stages, which reduce the resized window by a factor of 16 as the VQ encoder does.  It outputs a condition token $\mathbf{c}_n\in\mathbb{R}^{D}$ at every position $n$ of the $25\times25$ latent grid, and $C(\mathbf{x})$ denotes these 625 tokens.  Every latent position therefore has a condition token computed from the Sentinel-2 content around the same location.  A learned start token $\mathbf{s}\in\mathbb{R}^{D}$ follows the condition tokens, and the condition tokens and the start token form the prefix of the transformer input.

\subsection{Next-Scale Prediction}
\label{sec:nsp}
The next-scale transformer predicts the code maps $\mathbf{r}_1,\ldots,\mathbf{r}_K$ in $K$ steps.  Its input sequence consists of the prefix followed by one token map for each scale from the second on.  The token map of scale $k$ is the latent accumulated from the coarser code maps, downsampled to $s_k\times s_k$ and projected to $D$ channels,
\begin{equation}
    \mathbf{t}_k=W_{\mathrm{in}}\,\mathcal{A}_{s_k}\big(\mathbf{z}_{<k}\big)+\boldsymbol{\ell}_k,
    \label{eq:scale-input}
\end{equation}
where $\mathcal{A}_{s}$ is area downsampling to $s\times s$, $W_{\mathrm{in}}$ is a linear projection, and $\boldsymbol{\ell}_k$ is a learned embedding of scale $k$ added at every position.  The prefix receives learned position embeddings.  Each transformer block applies self-attention and a feed-forward layer, and the layer normalization before each of them takes its scale and shift from the start token.

All tokens are placed on the $25\times25$ latent grid.  Rotary position embedding rotates the queries and keys of every attention layer by the two-dimensional coordinates of a token, and position $(i,j)$ of an $s\times s$ map receives the coordinates $(25i/s,\,25j/s)$.  The condition tokens keep their grid coordinates, so relative positions in attention correspond to ground distances for the condition tokens and the tokens of all scales.

Attention follows the order of the scales.  A token of scale $k$ attends to the prefix and to all tokens of scales up to $k$, so every code map depends only on the input and on the coarser maps,
\begin{equation}
    p(\mathbf{r}_{1:K}\mid\mathbf{x})=
    \prod_{k=1}^{K}p_{\theta}\big(\mathbf{r}_{k}\mid\mathbf{r}_{<k},C(\mathbf{x})\big),
    \label{eq:factorization}
\end{equation}
which is the factorization of next-scale prediction \cite{tian2024var,qu2025varsr}.  The output at the start token gives the distribution of $\mathbf{r}_1$, and the output at each position of $\mathbf{t}_k$ gives the distribution of the code at the same position of $\mathbf{r}_k$, so all positions of a scale are predicted in parallel.  An output head with a layer normalization and a linear layer maps the final transformer feature $\mathbf{h}_{k,n}$ at position $n$ of scale $k$ to logits over the codebook.  Training uses teacher forcing, in which the reference code maps form $\mathbf{z}_{<k}$ in Eq.~(\ref{eq:scale-input}).  At inference, the transformer selects the most probable code at every position, the accumulated latent $\hat{\mathbf{z}}_{<k+1}=\hat{\mathbf{z}}_{<k}+\phi_k(\hat{\mathbf{r}}_k)$ takes the place of $\mathbf{z}_{<k+1}$ as the input of the next scale, and cached keys and values let each step process only the tokens of its own scale.

\subsection{Deterministic Residual Refinement}
\label{sec:drr}
The codes come from a finite codebook, so the accumulated latent $\hat{\mathbf{z}}_{<K+1}$ cannot match a continuous latent exactly.  After the last scale, the residual refiner estimates the remaining residual at every position $n$ of the $25\times25$ map.  Its target is
\begin{equation}
    \boldsymbol{\delta}_n=\big(\mathbf{z}^{*}-\mathbf{z}_{<K}\big)_n-\mathbf{e}\big(\hat{\mathbf{r}}_{K,n}\big),
    \label{eq:residual}
\end{equation}
the part of the latent at the last scale that the embedding of the code $\hat{\mathbf{r}}_{K,n}$ selected by the transformer does not represent.  The refiner is a small diffusion model applied to every position separately.  Its denoiser is a stack of residual MLP blocks, each modulated by the diffusion step and by the condition
\begin{equation}
    \mathbf{c}_n=\mathrm{LN}\big(\mathbf{h}_{K,n}+W_{\mathrm{in}}\,\mathbf{e}(\hat{\mathbf{r}}_{K,n})\big),
    \label{eq:refiner-condition}
\end{equation}
where $W_{\mathrm{in}}$ is the projection of Eq.~(\ref{eq:scale-input}) and $\mathrm{LN}$ denotes layer normalization, so the condition joins the transformer feature at $n$ with the embedding of the code selected there.  The refiner is trained with the diffusion loss $\mathcal{L}_{\mathrm{diff}}$ of improved denoising diffusion probabilistic models \cite{nichol2021improved}, and its estimate $\hat{\boldsymbol{\delta}}$ completes the predicted latent $\hat{\mathbf{z}}=\hat{\mathbf{z}}_{<K+1}+\hat{\boldsymbol{\delta}}$.

At inference, the refiner estimates the residual of each position with a short reverse diffusion process, and we omit the index $n$ below.  Starting from $\mathbf{u}_T\sim\mathcal{N}(\mathbf{0},\mathbf{I})$, each step computes
\begin{equation}
    \mathbf{u}_{t-1}=\boldsymbol{\mu}_{\theta}(\mathbf{u}_t,t,\mathbf{c})+\tau\,\sigma_t\,\boldsymbol{\epsilon}_t,\qquad
    \boldsymbol{\epsilon}_t\sim\mathcal{N}(\mathbf{0},\mathbf{I}),
    \label{eq:reverse}
\end{equation}
where $\sigma_t$ is the standard deviation of the reverse step, and standard sampling uses $\tau=1$.  The injected noise makes the estimate $\hat{\boldsymbol{\delta}}=\mathbf{u}_0$ a random draw whose variation is unrelated to the reference.  For a fixed condition, the expected squared error of such a draw decomposes as
\begin{equation}
    \mathbb{E}\big\|\boldsymbol{\delta}-\hat{\boldsymbol{\delta}}\big\|^2=
    \big\|\boldsymbol{\delta}-\mathbb{E}\hat{\boldsymbol{\delta}}\big\|^2+
    \mathbb{E}\big\|\hat{\boldsymbol{\delta}}-\mathbb{E}\hat{\boldsymbol{\delta}}\big\|^2,
    \label{eq:bias-variance}
\end{equation}
so the sampling variance adds error without moving the estimate toward the reference.  Deterministic residual refinement sets $\tau=0$ in Eq.~(\ref{eq:reverse}), so every reverse step follows the predicted mean.  The initial state $\mathbf{u}_T$ remains random, but its effect on the output becomes negligible, as quantified in Section~\ref{sec:ablation}.  The refinement reuses the trained refiner and its number of steps, so it requires no training and no additional computation.

\subsection{Training Objective}
\label{sec:objective}
The condition encoder, the next-scale transformer, and the residual refiner are trained jointly with the objective
\begin{equation}
    \mathcal{L}=\mathcal{L}_{\mathrm{CE}}+\lambda_{d}\,\mathcal{L}_{\mathrm{diff}}+\mathcal{L}_{1}+\lambda_{s}\,\mathcal{L}_{\mathrm{SSIM}},
    \label{eq:objective}
\end{equation}
while the VQ autoencoder stays frozen.  The code cross-entropy $\mathcal{L}_{\mathrm{CE}}$ is averaged over the positions of each scale, and the scales are weighted in proportion to their side length $s_k$.  The last code map holds about half of all 1,265 positions, so a plain average over positions would let it outweigh the coarser maps that set the tone and layout of the window.  The L1 loss $\mathcal{L}_{1}$ and the SSIM loss $\mathcal{L}_{\mathrm{SSIM}}$ compare the reference window with the decoded image $G(\mathbf{z}_{<K}+\tilde{\mathbf{e}}_K)$, where $\mathbf{z}_{<K}$ accumulates the reference code maps of the first $K-1$ scales and $\tilde{\mathbf{e}}_K$ is a straight-through estimate of the embeddings of the last code map.  In the forward pass, $\tilde{\mathbf{e}}_K$ takes the embedding of the most probable code at each position, and its gradient is that of the probability-weighted mean of the embeddings of the most probable codes.  The pixel losses thus train the logits of the last scale through the frozen decoder.

\subsection{Aligned Window Decoding of Complete Scenes}
\label{sec:wd}
Sections~\ref{sec:cond} to~\ref{sec:drr} reconstruct one window at the size of a training patch.  A complete $160\times160$ input requires a $1600\times1600$ output and a $100\times100$ latent field, whereas the model is trained on a $25\times25$ latent grid.  We therefore process the resized condition image in overlapping $400\times400$ windows.  A training patch covers $40\times40$ input pixels, so its corner lies both on the input grid, with a spacing of 10 output pixels, and on the latent grid, with a spacing of 16, and the two grids coincide every 80 output pixels.  A window that starts elsewhere places the input pixels at offsets from the latent positions that never occur in training.  We therefore start every window at a multiple of 80 pixels, which places the windows of a complete scene at 0, 320, 640, 960, and 1200 pixels along each axis.  An overlap of a quarter window would start them at 0, 304, 608, 912, and 1200 pixels, so three of every five window rows and columns would begin inside an input pixel.

Each window must also be decoded at its training size.  The VQ decoder follows the VQGAN design \cite{esser2021vqgan}, in which group normalization computes its statistics over all positions of a feature map and self-attention at the latent resolution weights all positions.  Both operations were trained on $25\times25$ latents.  Fusing the window latents and decoding the $100\times100$ field in one pass, as in latent tiling, therefore normalizes every window with statistics of the whole scene and lets each position attend to sixteen times as many positions as in training.  We instead decode every window on its own and blend the decoded windows,
\begin{equation}
 \hat{\mathbf{y}}(p)=
 \frac{\sum_{j:p\in\Omega_j}\mathbf{w}_j(p)\,G\big(\hat{\mathbf{z}}_j\big)(p)}
 {\sum_{j:p\in\Omega_j}\mathbf{w}_j(p)},
 \label{eq:window-decoding}
\end{equation}
where $p$ is an output pixel, $\Omega_j$ is the support of window $j$ on the output grid, $\hat{\mathbf{z}}_j$ is its refined latent, and $\mathbf{w}_j$ is a positive Gaussian weight map centered on the window.  Every window is thus predicted, refined, and decoded as a training patch.  Aligned window decoding changes only how complete scenes are tiled and decoded and requires no training.

\section{Experiments}
\label{sec:experiments}

\subsection{Implementation and Evaluation}
\method\ is trained on the 102,544 training patches with the objective of Eq.~(\ref{eq:objective}), where $\lambda_d=2$ and $\lambda_s=0.05$, and all of its parameters are initialized from released natural-image weights \cite{qu2025varsr}.  All results of \method\ and its ablations use the same trained weights.  Inference selects the most probable code at every position, uses $\tau=0$, and assembles complete scenes by aligned window decoding.

PSNR on RGB and SSIM \cite{wang2004ssim} on the luminance channel measure fidelity to the reference, LPIPS \cite{zhang2018lpips} measures perceptual distance, and FID \cite{heusel2017fid} compares the distributions of predictions and references.  All methods are evaluated with the same code on the 3,015 complete test scenes, without border cropping or output color correction.

\subsection{Comparison Methods}
EDSR-baseline \cite{lim2017edsr}, SwinIR \cite{liang2021swinir}, and TTST \cite{xiao2024ttst} represent reconstruction networks, and Thera \cite{becker2023thera} represents a continuous image field.  Real-ESRGAN+ \cite{wang2021realesrgan} and Satlas-SR \cite{wolters2023satlas} represent adversarial models, EDiffSR \cite{xiao2024ediffsr} and TexADiff \cite{zhang2026texadiff} represent diffusion models, and VARSR \cite{qu2025varsr} is the released natural-image model.  EDSR-baseline, SwinIR, TTST, Real-ESRGAN+, and EDiffSR are trained on the training partition with their outputs adapted to $\times10$, and Table~\ref{tab:training} lists the budgets of the first four.  TexADiff and VARSR are applied with their released weights.  RSDiffSR and image-only SGDM are not compared because a released implementation or image-only checkpoint was not available.

\begin{table}[!t]
\caption{Training Budgets of the Retrained Baselines}
\label{tab:training}
\centering
\setlength{\tabcolsep}{4pt}
\renewcommand{\arraystretch}{1.08}
\begin{tabular*}{\columnwidth}{@{\extracolsep{\fill}}lrr@{}}
\toprule
Method & Batch & Steps \\
\midrule
EDSR-baseline & 16 & 150k \\
SwinIR & 8 & 150k \\
TTST & 4 & 150k \\
Real-ESRGAN+ & 4 & 150k \\
\bottomrule
\end{tabular*}
\end{table}

\begin{table*}[!t]
\caption{Full-Scene $\times10$ Reconstruction on YREB-SR.  All Reported Scores Use the Same 3,015 Test Pairs.  Best and Second-Best Results Are Bold and Underlined.}
\label{tab:main}
\centering
\renewcommand{\arraystretch}{1.02}
\setlength{\tabcolsep}{5.2pt}
\begin{tabular*}{0.96\textwidth}{@{\extracolsep{\fill}}llcccc@{}}
\toprule
Method & Representation & PSNR (dB)$\uparrow$ & SSIM$\uparrow$ & LPIPS$\downarrow$ & FID$\downarrow$ \\
\midrule
\rowcolor{tablegroup}\multicolumn{6}{@{}l}{\textit{Reconstruction and continuous-representation methods}} \\
Bicubic & Interpolation & 14.5051 & 0.4626 & 0.9534 & 215.9498 \\
EDSR-baseline \cite{lim2017edsr} & CNN & 24.0986 & 0.6596 & 0.6831 & 68.2941 \\
SwinIR \cite{liang2021swinir} & Transformer & 24.5078 & 0.6706 & 0.6371 & 58.0257 \\
TTST \cite{xiao2024ttst} & RS transformer & 24.7157 & 0.6739 & 0.6201 & 48.9027 \\
Thera \cite{becker2023thera} & Neural field & \underline{25.2978} & \underline{0.6855} & 0.5679 & 37.8334 \\
\addlinespace[1pt]
\rowcolor{tablegroup}\multicolumn{6}{@{}l}{\textit{Perceptual and generative methods}} \\
Real-ESRGAN+ (paired) \cite{wang2021realesrgan} & GAN & 21.9581 & 0.6179 & 0.3269 & 47.2463 \\
Satlas-SR \cite{wolters2023satlas} & RS GAN & 23.2978 & 0.6328 & \underline{0.2150} & \underline{16.4924} \\
EDiffSR \cite{xiao2024ediffsr} & RS diffusion & 16.1011 & 0.4386 & 0.4529 & 99.6146 \\
TexADiff \cite{zhang2026texadiff} & Texture-aware diffusion & 14.8779 & 0.4332 & 0.8325 & 175.9166 \\
\addlinespace[1pt]
\rowcolor{tablegroup}\multicolumn{6}{@{}l}{\textit{Visual autoregressive methods}} \\
VARSR \cite{qu2025varsr} & Natural-image VAR & 14.1794 & 0.4381 & 0.6840 & 173.3056 \\
\midrule
\textbf{\method\ (ours)} & Cross-sensor VAR & \textbf{27.6008} & \textbf{0.7725} & \textbf{0.1451} & \textbf{7.1486} \\
\bottomrule
\end{tabular*}
\end{table*}

\subsection{Quantitative Results}
Table~\ref{tab:main} reports the four metrics on the 3,015 test scenes.  \method\ obtains 27.6008~dB PSNR, 0.7725 SSIM, 0.1451 LPIPS, and 7.1486 FID, the best value of every metric.  The compared methods fall into two groups.  The reconstruction networks and the continuous image field attain the highest fidelity among the baselines, led by Thera with 25.30~dB PSNR and 0.6855 SSIM, but their LPIPS of 0.57 to 0.68 and FID of 37.8 to 68.3 indicate smooth textures.  The adversarial models reverse this balance.  Satlas-SR lowers LPIPS to 0.2150 and FID to 16.49, while its PSNR falls 2.00~dB below that of Thera.  \method\ does not follow this trade-off.  It exceeds Thera by 2.30~dB PSNR and 0.087 SSIM and lowers the LPIPS and FID of Satlas-SR by 33\% and 57\%, respectively.

The diffusion models and the released VARSR stay close to bicubic interpolation.  EDiffSR reaches 16.10~dB, TexADiff applies its released $\times4$ weights to the $\times10$ grid and reaches 14.88~dB, and VARSR transferred from natural images reaches 14.18~dB with an FID of 173.31.  Priors learned from natural images or synthetic degradations therefore do not transfer to the cross-sensor $\times10$ task without training on real pairs.

Average scores can conceal scenes on which a method fails.  Table~\ref{tab:paired} therefore compares \method\ with every retrained baseline scene by scene.  \method\ attains a higher PSNR, a higher SSIM, and a lower LPIPS than each of these methods on every one of the 3,015 test scenes.  Against Thera, the PSNR gain has a median of 2.30~dB, a 5th percentile of 1.39~dB, and a minimum of 0.09~dB, and the 95\% bootstrap interval of the mean gain spans 2.28 to 2.32~dB.  Against Satlas-SR, the mean PSNR gain is 4.30~dB, and LPIPS is lower by 0.070.

\begin{table}[!t]
\caption{Scene-Wise Comparison of \method\ with the Retrained Baselines on the 3,015 Test Scenes.  Differences Are \method\ Minus Baseline, and the Last Column Gives the Share of Scenes on Which \method\ Is Better in All Three Metrics.}
\label{tab:paired}
\centering
\setlength{\tabcolsep}{4pt}
\renewcommand{\arraystretch}{1.05}
\begin{tabular*}{\columnwidth}{@{\extracolsep{\fill}}lcccc@{}}
\toprule
Baseline & $\Delta$PSNR (dB) & $\Delta$SSIM & $\Delta$LPIPS & Better (\%) \\
\midrule
EDSR-baseline & +3.50 & +0.113 & $-$0.538 & 100 \\
SwinIR & +3.09 & +0.102 & $-$0.492 & 100 \\
TTST & +2.89 & +0.099 & $-$0.475 & 100 \\
Thera & +2.30 & +0.087 & $-$0.423 & 100 \\
Real-ESRGAN+ & +5.64 & +0.155 & $-$0.182 & 100 \\
Satlas-SR & +4.30 & +0.140 & $-$0.070 & 100 \\
\bottomrule
\end{tabular*}
\end{table}

\subsection{Spectral, Complexity, and Regional Analysis}
\label{sec:analysis}
\begin{figure*}[!t]
    \centering
    \includegraphics[width=\textwidth]{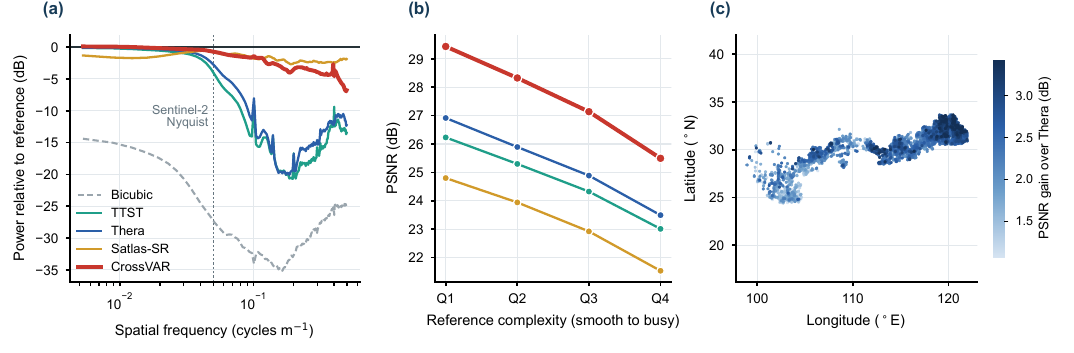}
    \caption{Scene-wise analysis on the 3,015 test scenes.  (a) Radially averaged power spectrum of the reconstructions relative to the Jilin-1 reference.  The dotted line marks the Nyquist frequency of the Sentinel-2 input.  (b) Mean PSNR in quartiles of the mean gradient magnitude of the reference.  (c) PSNR gain of \method\ over Thera at the scene centers.}
    \label{fig:analysis}
\end{figure*}

Fig.~\ref{fig:analysis} examines where the gain of \method\ arises.  Panel~(a) compares the radially averaged power spectrum of each reconstruction with that of the Jilin-1 reference.  Below 0.05 cycles per meter, the Nyquist frequency of the Sentinel-2 input, the reconstruction networks and Thera stay within 1.8~dB of the reference and \method\ within 0.2~dB, since this band is observed in the input.  Above it, the methods separate.  Between 0.05 and 0.15 cycles per meter, the reconstruction networks and Thera retain 10.6 to 14.0~dB less power than the reference, Satlas-SR 1.4~dB less, and \method\ 1.8~dB less.  The mean gradient magnitude of the reconstructions reaches 0.28 to 0.34 of that of the reference for the reconstruction networks and Thera, 0.77 for Satlas-SR, and 0.74 for \method.  \method\ therefore keeps nearly as much fine structure as the adversarial model while its PSNR is 4.30~dB higher.

Panel~(b) groups the test scenes into quartiles of the mean gradient magnitude of the reference, from the smoothest to the most textured quarter.  PSNR falls with complexity for every method, and \method\ remains the most accurate in each quartile.  Its gain over Thera decreases from 2.52~dB in the smoothest quartile to 2.00~dB in the most textured one, and its LPIPS rises from 0.116 to 0.173, compared with 0.493 to 0.632 for Thera and 0.179 to 0.250 for Satlas-SR.

Panel~(c) maps the PSNR gain over Thera at the scene centers.  The gain is positive across the whole basin.  It averages 1.91~dB in the upper reaches west of 111$^\circ$E (859 scenes), 2.36~dB in the middle reaches between 111$^\circ$E and 116$^\circ$E (490 scenes), and 2.49~dB in the lower reaches (1666 scenes).  The upper reaches also have the lowest PSNR for every method, 26.71~dB for \method\ against 27.99~dB in the lower reaches.

\begin{table*}[!t]
\caption{Full-Scene Efficiency.  Times Are Milliseconds per Scene and Memory Is Peak MiB.}
\label{tab:efficiency}
\centering
\renewcommand{\arraystretch}{1.0}
\setlength{\tabcolsep}{4pt}
\begin{tabular*}{0.92\textwidth}{@{\extracolsep{\fill}}lccccc@{}}
\toprule
Method & Parameters (M) & Precision & Model time & I/O-inclusive & Peak memory \\
\midrule
Bicubic & -- & CPU & 26.44 & 127.91 & -- \\
EDSR-baseline & 1.39 & FP32 & 2.92 & 384.07 & 124 \\
SwinIR & 11.99 & FP32 & 106.41 & 440.12 & 499 \\
TTST & 21.76 & FP32 & 189.63 & 503.57 & 1,632 \\
Thera & 5.85 & FP32 & 303.50 & 717.46 & 4,534 \\
Real-ESRGAN+ & 20.32 & FP32 & 44.96 & 446.78 & 2,283 \\
Satlas-SR & 16.73 & FP32 & 42.30 & 404.91 & 2,163 \\
EDiffSR & 20.40 & FP32 & 88,213.74 & 88,696.91 & 4,974 \\
TexADiff & 3,525.17 & FP16 & 9,766.84 & 11,621.83 & 14,588 \\
VARSR & 1,210.86$^{\dagger}$ & FP16 & 3,273.50 & 3,370.45 & 15,140 \\
\midrule
\method\ (ours) & 1,210.25$^{\dagger}$ & FP16 & 2,765.14 & 2,849.16 & 9,219 \\
\bottomrule
\end{tabular*}
\vspace{1mm}

\begin{minipage}{0.92\textwidth}
\footnotesize\textit{Note:} times are milliseconds per complete scene on one RTX~4090.  $^{\dagger}$Includes the frozen VQ autoencoder.
\end{minipage}
\end{table*}

\begin{figure*}[!t]
    \centering
    \includegraphics[width=\textwidth]{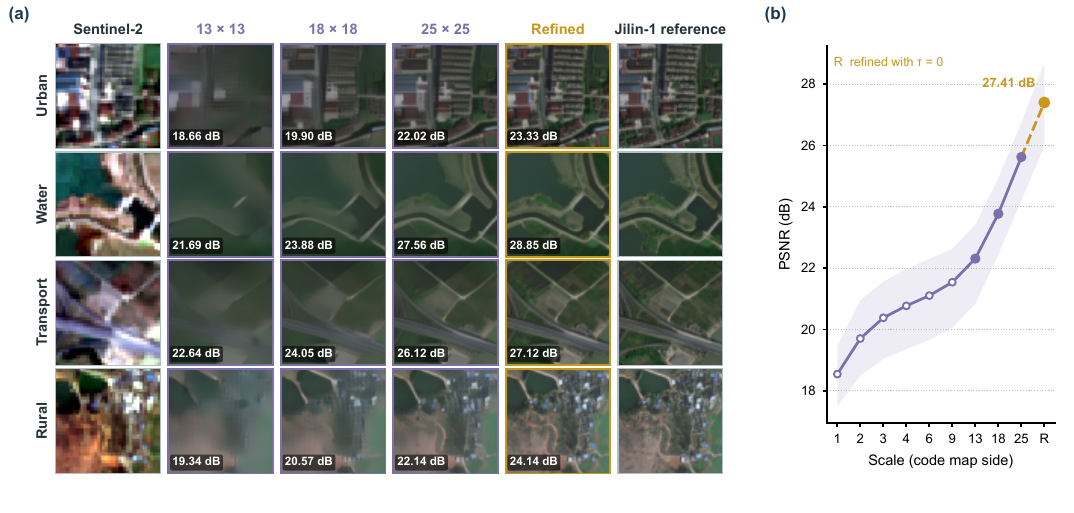}
    \caption{Coarse-to-fine reconstruction.  (a) Four test windows decoded after the $13\times13$, $18\times18$, and $25\times25$ code maps and after deterministic residual refinement, with PSNR against the Jilin-1 reference.  (b) Mean PSNR after every scale on 427 validation scenes, with the interquartile range shaded.}
    \label{fig:scales}
\end{figure*}

\begin{figure*}[!t]
    \centering
    \includegraphics[width=\textwidth]{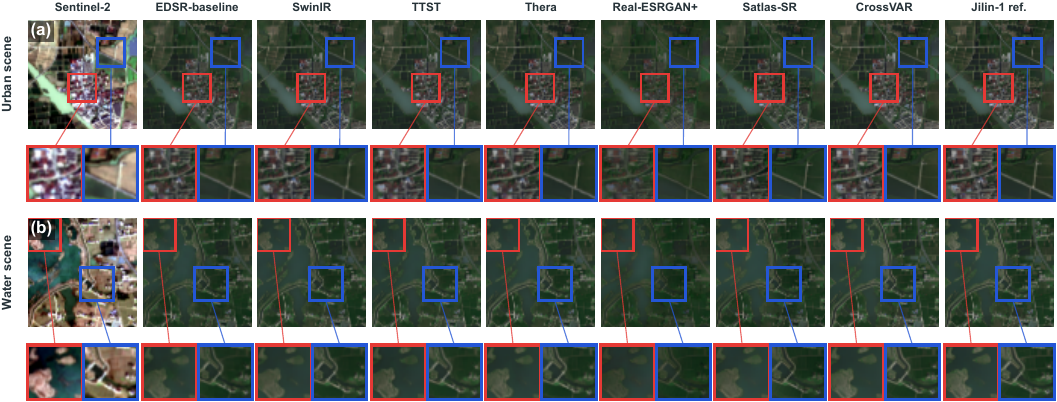}
    \vspace{2pt}
    \includegraphics[width=\textwidth]{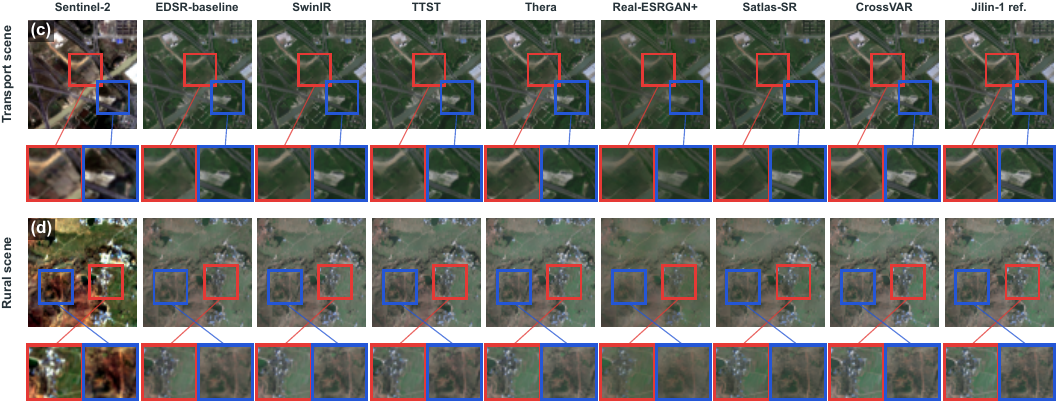}
    \caption{Complete-scene and local-detail comparisons on four test scenes.  Seven learned methods are shown between the Sentinel-2 input and the Jilin-1 reference, and red and blue boxes mark the enlarged regions.  Only the Sentinel-2 input is contrast-stretched for display.}
    \label{fig:qualitative}
\end{figure*}

\begin{figure}[!t]
    \centering
    \includegraphics[width=\columnwidth]{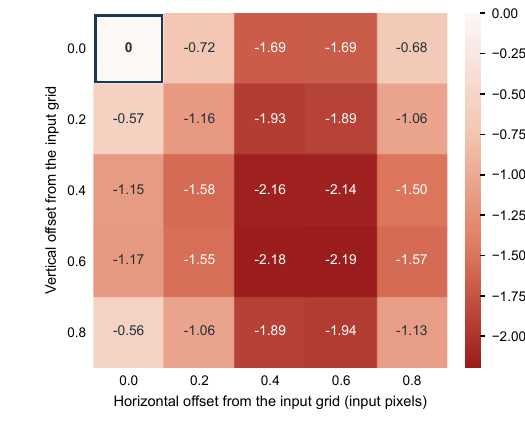}
    \caption{PSNR change of single windows on 427 validation scenes when the window start is moved away from the Sentinel-2 pixel grid.  The outlined cell is the aligned start.}
    \label{fig:phase}
\end{figure}

\subsection{Coarse-to-Fine Reconstruction}
\label{sec:scales}
Fig.~\ref{fig:scales} follows the reconstruction across the nine scales.  On the 427 validation scenes, the first six code maps set the overall tone and raise PSNR from 18.55 to 21.54~dB.  The layout of buildings, roads, and water appears at $13\times13$, and the $18\times18$ and $25\times25$ maps add the fine structure and raise PSNR to 25.63~dB.  Deterministic residual refinement adds a further 1.78~dB and sharpens the textures in all four examples.

\subsection{Qualitative Results}
Figure~\ref{fig:qualitative} compares four scenes with urban, water, transport, and rural content.  In the urban scene, the reconstruction networks and Thera merge neighboring buildings into blurred patches, whereas \method\ separates individual buildings and keeps the narrow field roads in the enlarged regions continuous.  In the water scene, the levees between ponds become faint in the outputs of the reconstruction networks, and \method\ reproduces them as thin continuous lines close to the reference.  Across the four scenes, the reconstruction networks produce smooth textures and the adversarial and autoregressive models restore sharper local structure, consistent with the spectra in Fig.~\ref{fig:analysis}(a) and with their lower LPIPS.

\subsection{Computational Cost}
Table~\ref{tab:efficiency} reports the cost of reconstructing one complete scene.  \method\ requires 2.77~s of model execution per scene, less than the released VARSR and the iterative diffusion models but more than the one-pass reconstruction networks.  Decoding each window on its own lowers the peak memory from 15,109~MiB to 9,219~MiB with 6\% more model time.

\subsection{Ablation Study}
\label{sec:ablation}
Table~\ref{tab:ablation} isolates the two components on the 3,015 test scenes.  All four settings use the same weights and evaluator.  The base model samples the residual with $\tau=1$, tiles the scene with an overlap of a quarter window, and decodes the fused scene latent in one pass.

\begin{table}[!t]
\caption{Ablation of Deterministic Residual Refinement (DRR) and Aligned Window Decoding (AWD) on the 3,015 Test Scenes}
\label{tab:ablation}
\centering
\setlength{\tabcolsep}{4pt}
\renewcommand{\arraystretch}{1.05}
\begin{tabular*}{\columnwidth}{@{\extracolsep{\fill}}cccccc@{}}
\toprule
DRR & AWD & PSNR$\uparrow$ & SSIM$\uparrow$ & LPIPS$\downarrow$ & FID$\downarrow$ \\
\midrule
 &  & 25.93 & 0.6960 & 0.1796 & 10.29 \\
\checkmark &  & 26.27 & 0.7076 & 0.1695 & 9.81 \\
 & \checkmark & 27.01 & 0.7509 & 0.1539 & \textbf{6.38} \\
\checkmark & \checkmark & \textbf{27.60} & \textbf{0.7725} & \textbf{0.1451} & 7.15 \\
\bottomrule
\end{tabular*}
\end{table}

Aligned window decoding improves all four metrics at both temperatures and raises PSNR by 1.08~dB with $\tau=1$ and by 1.33~dB with $\tau=0$, on every test scene.  Deterministic residual refinement raises PSNR by 0.34~dB and 0.60~dB and improves SSIM and LPIPS on every scene, while FID prefers the sampled residual once the windows are aligned \cite{blau2018perception}.  Together, the two components raise PSNR from 25.93 to 27.60~dB and lower FID from 10.29 to 7.15.

Deterministic refinement also makes the output nearly independent of the random seed.  On 64 validation patches, two runs with different seeds agree to 52.5~dB PSNR with $\tau=0$, against 31.5~dB with $\tau=1$.

Table~\ref{tab:fusion} compares where the windows start, where they are fused, and at which size the decoder runs.  Latent fusion blends the window latents before decoding, and RGB fusion blends windows decoded on their own as in Eq.~(\ref{eq:window-decoding}).  Starting the windows on the input grid gives most of the gain, 0.99 to 1.15~dB for every fusion rule.  Decoding each window at the training size adds a further 0.31~dB, and RGB fusion of the decoded windows is best on all four metrics.  This assembly was selected on 427 complete validation scenes, formed from the 16 patches of every fifth validation scene, where it reached 27.64~dB against 26.32~dB for latent fusion with one-pass decoding.

\begin{table}[!t]
\caption{Complete-Scene Assembly Rules with $\tau=0$ on the 3,015 Test Scenes}
\label{tab:fusion}
\centering
\setlength{\tabcolsep}{3pt}
\renewcommand{\arraystretch}{1.05}
\begin{tabular*}{\columnwidth}{@{\extracolsep{\fill}}lllcccc@{}}
\toprule
Windows & Fusion & Decoding & PSNR$\uparrow$ & SSIM$\uparrow$ & LPIPS$\downarrow$ & FID$\downarrow$ \\
\midrule
On grid & None & Scene & 26.39 & 0.7406 & 0.1804 & 10.98 \\
Off grid & Latent & Scene & 26.27 & 0.7076 & 0.1695 & 9.81 \\
Off grid & Latent & Window & 26.51 & 0.7145 & 0.1595 & 8.94 \\
Off grid & RGB & Window & 26.45 & 0.7131 & 0.1558 & 7.96 \\
On grid & Latent & Scene & 27.26 & 0.7645 & 0.1609 & 9.08 \\
On grid & Latent & Window & 27.58 & 0.7708 & 0.1506 & 8.21 \\
On grid & RGB & Window & \textbf{27.60} & \textbf{0.7725} & \textbf{0.1451} & \textbf{7.15} \\
\bottomrule
\end{tabular*}
\end{table}

Fig.~\ref{fig:phase} tests the window start directly on the 427 validation scenes.  Single windows were started 0.2 to 0.8 input pixels away from the grid in each axis and decoded on their own.  PSNR falls steadily as the start moves away from the nearest input-pixel boundary, by 0.6 to 0.7~dB for a shift of 0.2 input pixels in one axis and by up to 2.2~dB when both axes start near the middle of an input pixel, and the aligned start is best in 87\% of the scenes.

\section{Discussion}
\label{sec:discussion}

\subsection{Fidelity and Perceptual Quality at a Large Magnification}
Distortion and perceptual quality are known to compete in SR \cite{blau2018perception}, and the baselines in Table~\ref{tab:main} follow this pattern.  At $\times10$, the competition is governed by the frequencies that the Sentinel-2 input does not observe.  A network trained with a pixel loss predicts the conditional mean of these frequencies, and because many fine structures are consistent with the same input pixels, this mean carries little power, as Fig.~\ref{fig:analysis}(a) shows.  An adversarial model restores the power, but the restored structures need not lie where the reference places them, which lowers PSNR.  \method\ selects the fine structure as discrete codes conditioned on coarser code maps, and the coarse maps fix the layout that the input constrains (Fig.~\ref{fig:scales}).  A likely explanation for its gain in both fidelity and perceptual quality is that the fine codes are placed relative to an already correct layout, so the reconstruction keeps both the power and much of the position of fine structure.  The residual temperature offers an explicit control over the remaining balance.  With aligned windows, $\tau=1$ gives the lowest FID and $\tau=0$ the best PSNR, SSIM, and LPIPS (Table~\ref{tab:ablation}), and $\tau=0$ is used throughout because the benchmark measures agreement with the reference.

\subsection{Window Alignment in Tiled Reconstruction}
The misalignment addressed in Section~\ref{sec:wd} arises whenever the number of output pixels per input pixel does not divide the latent stride.  With a stride of 16, every window on the latent grid also starts on the input grid at $\times2$, $\times4$, $\times8$, and $\times16$, which is a probable reason why tiling at these common factors has not exposed the effect.  At $\times3$ and $\times6$ the two grids coincide every 48 output pixels, and at $\times10$ every 80.  Fig.~\ref{fig:phase} shows that a start 0.2 input pixels away from the grid already costs 0.6 to 0.7~dB.  Latent generative SR models that tile complete scenes at such factors may therefore benefit from starting their windows on the common grid and from decoding each window at the size used in training.

\subsection{Limitations}
The Sentinel-2 composites and the Jilin-1 references come from different sensors and different acquisition dates, and their footprints are matched geographically rather than registered pixel by pixel.  Part of the measured error therefore reflects sensor and temporal differences that no reconstruction from the input can remove, and the absolute values of the reference-based metrics should be read with this in mind.

Structures smaller than a Sentinel-2 pixel, such as individual buildings and narrow field boundaries, are predicted from the learned prior.  These structures are not observed in the input, so the reconstructions are suited to visual interpretation and to mapping at the level of the observed layout rather than to measuring individual objects below 10~m.

The benchmark covers the Yangtze River Economic Belt, and its random scene-level partition is not a geographically blocked holdout.  Transfer to other climates and landscapes and the use of other spectral bands remain to be tested.  \method\ also requires 2.77~s and 9.2~GB per complete scene with about 1.2 billion parameters, much more than the compact reconstruction networks.  Finally, on validation patches, taking the code maps of the first eight scales from the reference raises PSNR from 27.36 to 29.23~dB, whereas decoding the unquantized reference latent reaches 32.45~dB, so most of the remaining error lies in the last scale and the continuous residual.

\section{Conclusion}
\label{sec:conclusion}
We studied real-world $\times10$ reconstruction from a single Sentinel-2 RGB composite using the 11,559 geographically matched pairs in \dataset.  \method\ predicts the 1-m latent as nine code maps and a continuous residual, refines the residual without injected noise, and reconstructs complete scenes in windows that keep the input alignment and the decoder size of training.  On the fixed 3,015-scene test split, \method\ obtains 27.60\ dB PSNR, 0.7725\ SSIM, 0.1451\ LPIPS, and 7.15\ FID.  Scene-wise and spectral analyses show that the advantage over the retrained baselines holds across scene complexity and region and that \method\ keeps most of the power of fine structure above the Sentinel-2 Nyquist frequency while placing it far more accurately than adversarial models.  Future work will estimate the continuous residual more closely from the Sentinel-2 input, add multi-temporal and multi-spectral observations, and reduce the inference cost.

\section*{CRediT Authorship Contribution Statement}
Tailai Chen and Yuluxin Fu: Writing -- original draft, Software, Conceptualization, Formal analysis. Shu Luo: Data curation, Resources. \rev{Jie Deng, Yanan Wu and Danfeng Hong: Supervision, Conceptualization, Writing -- review \& editing, Resources, Funding acquisition.}

\section*{Declaration of Competing Interest}
The authors declare that they have no known competing financial interests or personal relationships that could have appeared to influence the work reported in this paper.

\section*{Funding}
This research was supported by the National Natural Science Foundation of China (Grant No. 32402311).

\section*{Acknowledgment}
\rev{The authors thank all collaborators and institutions involved in this research for their valuable contributions and assistance. All authors approved the final version of the manuscript.}

\IEEEtriggeratref{25}
\bibliographystyle{IEEEtran}
\bibliography{references}
\newpage
\begin{IEEEbiography}[{\includegraphics[width=1in,height=1.25in,clip,keepaspectratio]{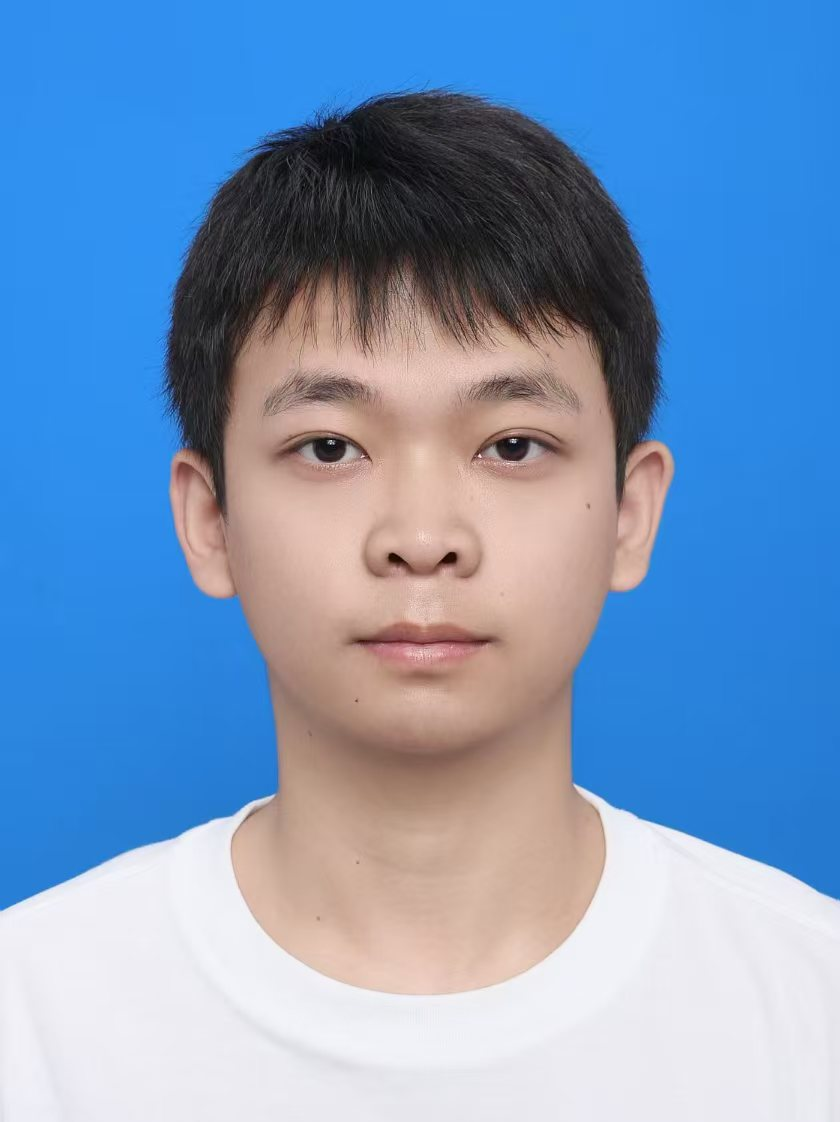}}]{Tailai Chen} is currently pursuing the Ph.D. degree with the Institute of Automation, Chinese Academy of Sciences, Beijing, China. His research interests include computer vision, open-world learning, multimodal learning, and scientific machine learning.
\end{IEEEbiography}
\vskip -2\baselineskip plus -1fil

\begin{IEEEbiography}[{\includegraphics[width=1in,height=1.25in,clip,keepaspectratio]{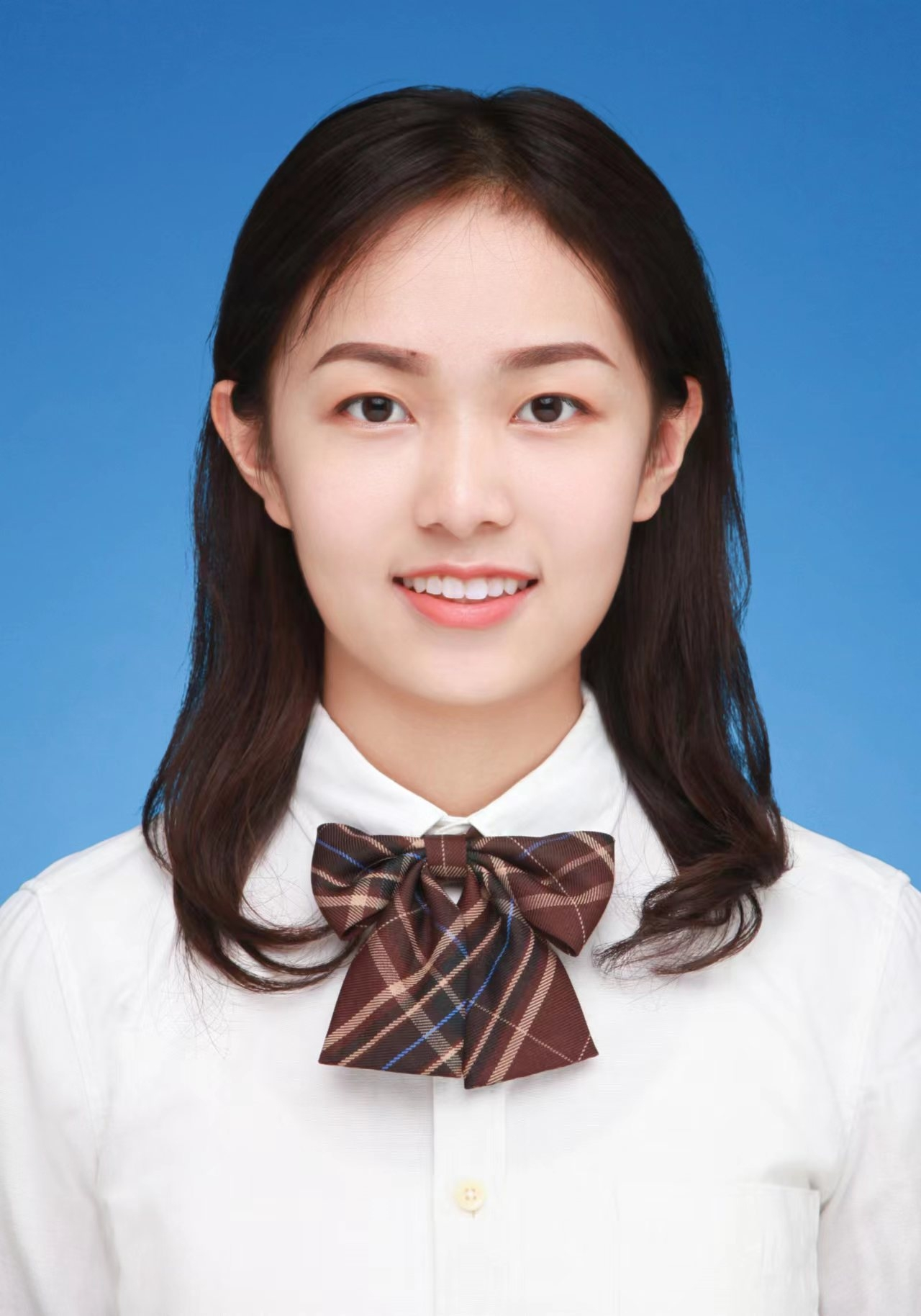}}]{Yuluxin Fu} received the B.S. degree in Artificial Intelligence from the College of Information and Electrical Engineering, China Agricultural University, Beijing, China. She is currently pursuing the Ph.D. degree with the Aerospace Information Research Institute, Chinese Academy of Sciences, Beijing, China. Her research interests include remote sensing vision foundation models and fine-grained water body extraction.
\end{IEEEbiography}
\vskip -2\baselineskip plus -1fil

\begin{IEEEbiography}[{\includegraphics[width=1in,height=1.25in,clip,keepaspectratio]{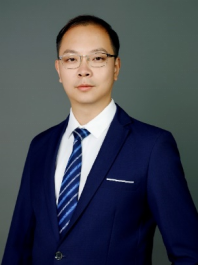}}]{Jie Deng}
received the Ph.D. degree in plant pathology from China Agricultural University, Beijing, China, in 2023. He was a Postdoctoral Researcher with the Aerospace Information Research Institute, Chinese Academy of Sciences, from 2023 to 2025. He is currently an Associate Professor with China Agricultural University. His research interests include agricultural remote sensing, crop disease monitoring, and artificial intelligence.
\end{IEEEbiography}
\vskip -2\baselineskip plus -1fil

\begin{IEEEbiography}[{\includegraphics[width=1in,height=1.25in,clip,keepaspectratio]{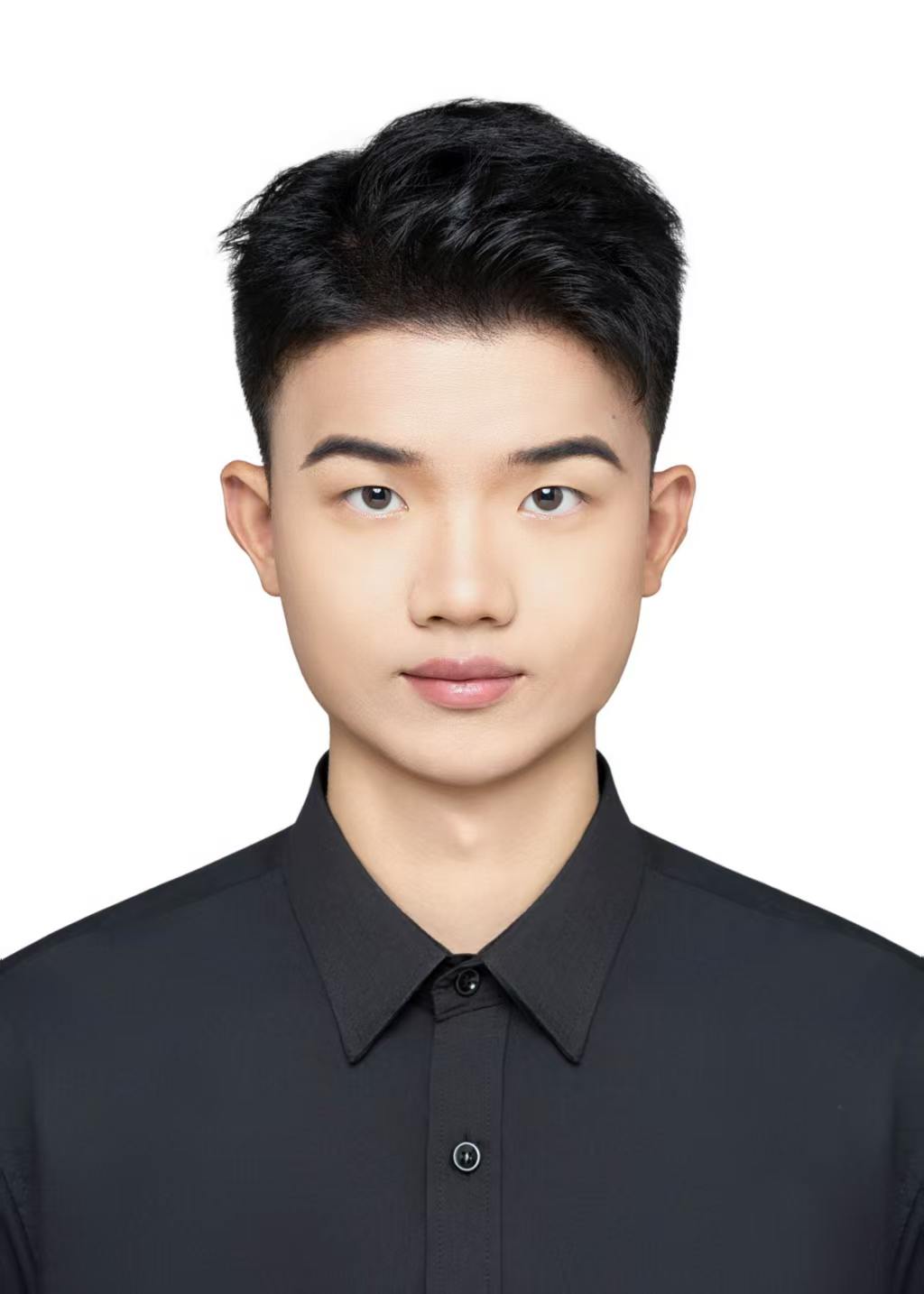}}]{Shu Luo}
received the B.E. degree from Jiangsu University, Zhenjiang, China. He is currently pursuing the M.S. degree in Agricultural Engineering and Information Technology at Beijing Forestry University, Beijing, China. His research interests include forestry remote sensing and deep learning.
\end{IEEEbiography}
\vskip -2\baselineskip plus -1fil

\begin{IEEEbiography}[{\includegraphics[width=1in,height=1.25in,clip,keepaspectratio]{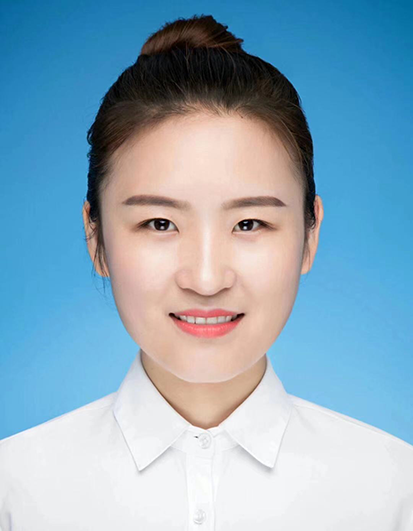}}]{Yanan Wu} received the Ph.D. degree from the School of Computer Science and Technology, Beijing Jiaotong University, Beijing, China, in 2024. She conducted postdoctoral research with the School of Computer Science, Carleton University, Ottawa, Canada, from 2024 to 2025. She is currently a Lecturer with the College of Information and Electrical Engineering, China Agricultural University, Beijing, China. Her research interests include open-world multi-modal information processing, data-driven foundation model adaptation, and smart agriculture. She has published more than 20 peer-reviewed papers in prestigious international conferences and journals, including CVPR, ICCV, ICLR, AAAI, ECCV, IJCAI, ACM Multimedia, and IEEE Transactions on Circuits and Systems for Video Technology, etc. She was also a Visiting Scholar with the Department of Computer Science and Software Engineering, Concordia University, Montreal, Canada, in 2023.
\end{IEEEbiography}
\vskip -2\baselineskip plus -1fil

\begin{IEEEbiography}[{\includegraphics[width=1in,height=1.25in,clip,keepaspectratio]{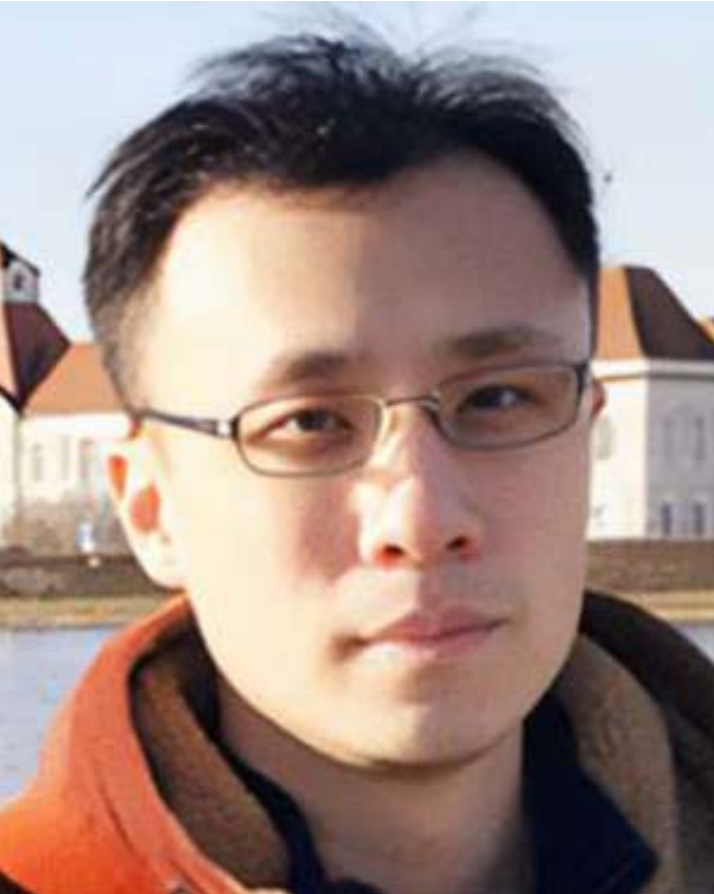}}]{Danfeng Hong}
(Senior Member, IEEE) received the Dr.-Ing. degree (summa cum laude) in signal and information processing from the Signal Processing in Earth Observation (SiPEO), Technical University of Munich, Munich, Germany, in 2019. He is currently a Full Professor with Southeast University, Nanjing, China, and previously served as a Full Professor with the Aerospace Information Research Institute, Chinese Academy of Sciences, Beijing, China. His research interests include artificial intelligence, multimodal perception, foundation models, hyperspectral remote sensing, and Earth science. Dr. Hong serves as an Associate Editor for IEEE TRANSACTIONS ON PATTERN ANALYSIS AND MACHINE INTELLIGENCE (TPAMI), IEEE TRANSACTIONS ON IMAGE PROCESSING (TIP), and IEEE TRANSACTIONS ON GEOSCIENCE AND REMOTE SENSING (TGRS). He has been recognized as a Highly Cited Researcher by Clarivate Analytics since 2022.
\end{IEEEbiography}

\end{document}